\documentclass{article}

\usepackage{hyperref}
\usepackage{url}
\usepackage{graphicx}
\usepackage{algpseudocode}
\usepackage{algorithm}
\usepackage{cite}
\usepackage{amsmath}
\usepackage{authblk}

\newcommand{\Wout}{\textbf{W}_{out}}
\newcommand{\rinit}{\textbf{r}_0}
\newcommand{\xinit}{\textbf{x}_0}
\newcommand{\param}{\textbf{v}}
\newcommand{\testsig}{\textbf{s}}
\newcommand{\latent}{\textbf{e}}
\newcommand{\library}{\textbf{L}}
\newcommand{\librarymember}{\textbf{l}^i}

\title{A Meta-learning Approach to Reservoir Computing: Time Series Prediction with Limited Data}

\author[1]{Daniel Canaday}
\author[1]{Andrew Pomerance}
\author[2]{Michelle Girvan}
\affil[1]{Potomac Research, LLC, Alexandria, VA 22314, USA}
\affil[2]{Department of Physics, University of Maryland, College Park, Maryland, 20742, USA}

\begin{document}

\maketitle

\begin{abstract}
Recent research has established the effectiveness of machine learning for data-driven prediction of the future evolution of unknown dynamical systems, including chaotic systems. However, these approaches require large amounts of measured time series data from the process to be predicted. When only limited data is available, forecasters are forced to impose significant model structure that may or may not accurately represent the process of interest. In this work, we present a Meta-learning Approach to Reservoir Computing (MARC), a data-driven approach to automatically extract an appropriate model structure from experimentally observed ``related'' processes that can be used to vastly reduce the amount of data required to successfully train a predictive model. We demonstrate our approach on a simple benchmark problem, where it beats the state of the art meta-learning techniques, as well as a challenging chaotic problem.
\end{abstract}

\section{Introduction}\label{Introduction}

Time series prediction is fundamentally and practically important across a wide variety of domains, and approaches to this task vary wildly in complexity and sophistication.  Some examples include parameter estimation based on physical models \cite{bjornstad2002dynamics}, filtering techniques \cite{musoff2009fundamentals, joo2015time}, autoregressive-moving average models and their variants \cite{pappas2008electricity, holan2010arma, faruk2010hybrid}, and other machine learning models such as support vector machines \cite{kim2003financial}, deep learning \cite{yan2018financial, han2019review}, and recurrent neural networks \cite{han2004prediction, gers2002applying, pathak2018model}. 

Generally speaking, there is a trade-off inherent to these techniques: either significant structure needs to be imposed by a physical or mathematical model, or large amounts of training data is required for machine learning-based methods.  
We are interested in this work in predicting time series with extremely limited data and poor theoretical models. ``Extremely limited'' can be defined as the case where not enough information is available to reduce our uncertainty about the true dynamics that generate the time series; this, naturally, forces us to impose some structure.  Since our theoretical understanding is poor, we would like to automate an inductive process that learns the appropriate structure from other, related time series.  This is an example of knowledge transference, which is concerned with developing techniques to efficiently adapt solutions to similar problems and has received considerable interest in a variety of contexts \cite{lemke2015metalearning, feurer2015initializing, finn2017model, finn2019online}, including time series prediction \cite{oreshkin2020meta, talagala2018meta, lemke2010meta}.   This problem and potential solutions are often referred to as meta-learning or ``learning about learning'' \cite{olier2018meta, gaudet2020terminal, abbasi2012metafraud, kanda2012meta}. 

\begin{figure}[h]\label{fig:diagram}
\begin{center}
\includegraphics[width=\textwidth]{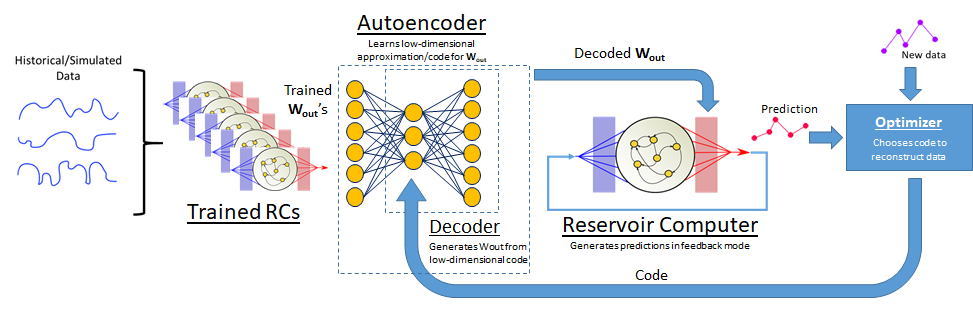}
\end{center}
\caption{A schematic illustration of the MARC algorithm. At left, a library of related time series (possibly from a combination of measurements and approximate simulated processes) is used to train an associated library of RC models. These RC features are used to train an autoencoder with a low-dimensional hidden layer. At right, the decoder portion of the autoencoder is used to map latent variables onto RC features, which then generates a time series. The latent variables are optimized such that this generated time series best agrees with a signal consisting of a limited amount of time series data to be predicted.}

\end{figure}

In this work, we introduce Meta-learning Approach to Reservoir Computing (MARC), a novel meta-learning technique whose goal is to enable accurate time series forecasting with vastly reduced training data.  Instead of imposing domain-specific model constraints, MARC learns appropriate model structure from a library of related examples in an automated fashion, illustrated schematically in Fig. \ref{fig:diagram}.  There are two distinct learning phases. In the first learning phase, a recurrent neural network model known as a reservoir computer (RC) is trained separately on each available time series in the library, yielding an associated predictive model.   While the RC model is an extremely flexible time series modelling framework, a relatively large number ($10^2-10^4$) of parameters must be learned to get satisfactory results, which increases the amount of training data required.  Although the number of RC parameters is large, we hypothesize that the trained RC models actually reside on a manifold in the RC parameter space of much lower dimensionality by virtue of the relatedness of the library processes.  The second learning phase, therefore, uses a feedforward deep autoencoder neural network \cite{lecun2015deep, kingma2013auto, tschannen2018recent} to project the low-dimensional manifold of learned RC feature vectors onto a relatively small set of latent representation parameters that encode the predictive models.  After the latent representation is learned, we can search this space to identify candidate RC feature vectors that explain new, incoming data. By constraining our attention to this reduced manifold of relevant RC features, we drastically reduce data requirements while avoiding overfitting because we are searching within the space of \textit{reasonable} RC models rather than the space of \textit{all} RC models.

An illustrative example that will be revisited in Sec. \ref{subsec:sine} is that of modeling a sine wave from only a few points on the curve.  While this is a simple problem, it has been previously used by others for algorithm validation and thus allows quantitative comparison of our algorithm to state-of-the-art few-shot meta-learning algorithms \cite{wu2018meta, finn2017model}.  In this toy regression problem, there are two parameters that define the time series--the amplitude and phase. If we know that the signal to be predicted is sinusoidal, very few noiseless observations are required.  However, if we were to use a standard machine learning approach without any prior knowledge of the signal's character, we would need significantly more observations to generate a usable predictive model.  This is, in essence, the idea behind MARC: rather than searching for the best fit among \textit{all} RC models, we search only on the manifold of RC models that approximate sinusoidal dynamics.  Instead of explicitly fitting to a sinusoid, MARC accomplishes this in a model-free manner by learning (via the autoencoder) the space of sinusoidal RCs from other examples of sinusoidal dynamics.

Our proposed approach stands in contrast to previous meta-learning approaches to time-series prediction problems. Many state-of-the-art meta-learning algorithms, such as MAML (\cite{finn2017model}), HSML (\cite{yao2019hierarchically}), LEO (\cite{rusu2018meta}) and others, use meta-knowledge to constrain initial parameters (or sets of initial parameters) from which a regressive model can be fine-tuned to a new task. By contrast, MARC uses meta-knowledge (the library) to constrain the \textit{complete} set of parameters for a predictive model. This is possible because, as we discuss in the following sections (see, in particular, Fig. \ref{fig:sineexample}), MARC creates a latent representation of the unknown degrees of freedom that distinguish library members from one another. Effectively, this allows us to do curve-fitting without ever accessing the true unknown parameters--rather, we fit within the latent space, from which the entire distribution of time series is represented.

The rest of this paper is organized as follows.  In Sec. \ref{sec:definitions}, we describe the scope of the problem space, including defining ``related time series,'' and quantifying the data requirements. In Sec. \ref{sec:algorithm}, we describe the three parts of the MARC framework: the RC model for time series prediction, compressing the resulting RC models with autoencoders, and then searching the latent variable space for usable predictive models with limited data.  In Sec. \ref{sec:results}, we demonstrate the practical application of the proposed algorithm with a number of simulated systems. Finally, we conclude with observations and considerations for further research in Sec. \ref{sec:conclusions}.

\section{Definitions and Scope}\label{sec:definitions}

We assume that knowledge of a class of processes is available in the form of a library of time series. We define a \textit{library} $\library$ as a set of \textit{members} $\librarymember$, where each $\librarymember$ is a set of time-labeled observations of the state of a dynamical system. That is, $\library=\{\librarymember\}$, and, as an example, each $\librarymember$ might be a discrete, evenly-spaced time sampling of a finite duration, continuous-time, state observation $\textbf{y}^i(t)$ from a library process $i$, where the library process $i$ and the process generating the signal to be predicted are assumed to be related. Although it is not necessary for our MARC technique, it is conceptually useful to consider the relatedness of the library members as modeled as an ensemble of systems where each $\textbf{y}^i(t)$ obeys
\begin{eqnarray}\label{eqnarray:dynamics}
    \dot{\textbf{x}}^i&=&f(\textbf{x}^i; \textbf{v}^i) \nonumber \\
    \textbf{y}^i&=&g(\textbf{x}^i; \textbf{v}^i)
\end{eqnarray}
subject to the initial condition $\textbf{x}^i(0) = \textbf{x}^i_0$ and an unobserved parameter vector $\textbf{v}^i$. We call $f$ and $g$ the \textit{dynamical} and \textit{observation} functions, respectively, and $\textbf{x}^i$ and $\textbf{y}^i$ the \textit{state-space} and \textit{observation} variables, respectively. With these definitions, each $\librarymember$ is a function of a particular initial condition $\textbf{x}_0^i$ and parameter vector $\textbf{v}^i$. Considering $f$ and $g$ to be fixed but unknown, the library members can be thought of as instances from an ensemble of realizations of a random process characterized by a probability distribution $p(\textbf{x}_0^i, \textbf{v}_0^i)$. In what follows, our discussion will be in the context of relatedness as given by Eq. \ref{eqnarray:dynamics}.

We call the signal to be predicted the test signal and denote it $\testsig(t)$. We assume that $\testsig(t)$ originates from a dynamical system given by Eq. \ref{eqnarray:dynamics} with initial condition $\xinit^s$ and parameter vector $\param^s$. A test signal is then only distinguished from a library member in that it is much shorter, and its parameter vector and/or initial condition may be outside of the support of $p(\textbf{x}_0^i, \textbf{v}_0^i)$, thus requiring any algorithm that learns about $\library$ to extrapolate.

There are several cardinalities that are relevant to these definitions, namely $\dim(\textbf{x)}$, $\dim(\textbf{y})$, $\dim(\param)$, $|\library|$, $|\librarymember|$, and $|\testsig|$, which are the dimension of the state space, the dimension of the observation space, the number of parameters, the number of library members, the number of observation time-points per library member, and the number of observations in the test signal, respectively. We consider an interesting regime to be where $|\testsig| \ll |\librarymember|$. More exactly, we consider a regime where each $\librarymember$ is sufficiently long that a useful predictive model can be obtained, but the same is not true for $\testsig$, hence the need to incorporate the knowledge from $\library$. On the other hand, we must also have $|\testsig|\dim(\textbf{y}) > \dim(\textbf{x}) + \dim(\textbf{v})$, otherwise the problem is underdetermined even if $\library$ were to provide complete information about $f$ and $g$.

Although we aim to be concrete and concise in our above definitions of ``relatedness'', we believe the applicability of MARC is quite broad. Reservoir computing methods are capable of representing a wide range of dynamical phenomena beyond what can be described by Eq. 1, including noisy dynamics (\cite{zimmermann2018observing}), nonautonomous systems (\cite{jaeger2003advances, patel2021using}), control systems (\cite{salmen2005echo, park2016online}), discrete-time maps (\cite{patel2021using}), partial differential equations (\cite{pathak2018model}), delay differential equations (\cite{jaeger2001echo}), and more. Generalizations of the version of MARC presented here are straightforward and only involve modifying the form of RC used to represent library members.  Moreover, MARC can handle multi-modal situations in which the library members have different functional forms (Appendix E). 

\section{The MARC Framework}\label{sec:algorithm}
In this section, we describe the MARC algorithm (summarized in Algorithm 1) and framework in more complete detail. As discussed above, the key idea is to 1) obtain RC features corresponding to an accurate predictive model for each library member, 2) train a deep autoencoder to learn a latent representation of these RC features, and then 3) fit, within latent variable space, a set of decoded RC features that best explains a small sample of new data.

\begin{algorithm}
\caption{MARC algorithm}\label{alg:cap}
\begin{algorithmic}[1]
\Require Library $\library = \{ \librarymember \}$, set of time series related to the process of interest
\Require each $\librarymember \in \library$ is long enough for successful RC training
\Require $T$, RC training procedure
\Require $R$, RC prediction procedure
\Require Short test signal $\testsig$
\State Choose RC training hyperparameters and time scale $c$
\State Randomly assign $\textbf{W}$, $\textbf{W}_{in}$, and $\textbf{b}$
\ForAll {$\librarymember \in \library$}
    \State $(\Wout^i, \rinit^i) \gets T(\librarymember; \textbf{W}$, $\textbf{W}_{in}$, $\textbf{b}, c)$
\EndFor
\State Let $A=D \circ E$ be the autoencoder
\State initialize autoencoder parameters $\Theta$ and reconstruction error
\While{reconstruction error too large}
    \State update parameters $\Theta$
    \State error $\gets |(\Wout^i, \rinit^i) - A(\Wout^i, \rinit^i)|^2$
\EndWhile
\State Initialize latent variable $\textbf{e}$ and prediction error 
\While {prediction error too large}
    \State update $\textbf{e}$
    \State error $\gets |R(D(\textbf{e}); \textbf{W}$, $\textbf{W}_{in}$, $\textbf{b}, c) - \testsig|^2$
\EndWhile
\State $(\Wout^s, \rinit^s) \gets D(\textbf{e})$
\State Generate predictions using $\Wout^s, \rinit^s$
\end{algorithmic}
\end{algorithm}

\subsection{Learning Time Series Representations with Reservoir Computing}\label{subsec:rc}
The first step in the MARC algorithm is to learn independent representations of each of the available time series in the library.  For this purpose, we use a type of recurrent neural network known as a reservoir computer (RC), a class of machine learning techniques often applied to processing temporal data \cite{jaeger2001echo, maass2002real, verstraeten2007experimental}.  We refer to the appendix for details on the RC model used in this work and \cite{vlachas2019forecasting} for a comparison to deep learning techniques for time series processing.  

The general RC procedure is as follows. Given an input time series $\textbf{u}(t)$, a recurrent neural network known as the \textit{reservoir} is driven with $\textbf{u}(t)$ and evolves according to
\begin{equation}\label{Reservoir Dynamics}
    c\dot{\textbf{r}}(t) = -\textbf{r}(t) + \textbf{tanh}\left(\textbf{W}\textbf{r}(t) + \textbf{W}_{in} \textbf{u}(t) + \textbf{b} \right)
\end{equation}
where $\textbf{r}(t)$ is the state of the reservoir at time $t$, $c$ is a time constant, $\textbf{W}$ is a \textit{fixed} matrix of recurrent connections, $\textbf{W}_{in}$ is a \textit{fixed} input matrix, and $\textbf{b}$ is a \textit{fixed} bias vector.  After an appropriate warm-up time, the state $\textbf{r}(t)$ of the reservoir is recorded and ridge regression is used to identify a time-independent, linear map such that $\textbf{W}_{out}\textbf{r}(t) \approx \textbf{o}^d(t)$, where $\textbf{o}^d(t)$ is a desired output time series.  This procedure, that maps an input time series $\textbf{u}$ to an initial state $\rinit$ and output weight matrix $\Wout$ is the procedure $T(\textbf{u}; \textbf{W}$, $\textbf{W}_{in}$, $\textbf{b}, c)$ in Algorithm 1.  The training procedure has several hyperparameters that are discussed in Appendix \ref{Reservoir Computing}.

For forecasting problems, the desired output is the same as the input signal, $\textbf{o}^d(t) = \textbf{u}(t)$. After the training period, future values of the input signal are predicted by replacing the input in Eq. 2 with the predicted output $\textbf{W}_{out}\textbf{r}$. This forms an autonomous predictive model with dynamics governed by
\begin{eqnarray}\label{eqnarray:closedloop}
    \dot{\textbf{r}}=f_{RC}(\textbf{r}; \textbf{W}_{out}) &=& -\textbf{r}/c + \tanh\left(\textbf{W}\textbf{r} + \textbf{W}_{in} \textbf{W}_{out} \textbf{r} + \textbf{b} \right)/c \nonumber \\
    \textbf{o}=g_{RC}(\textbf{r}; \textbf{W}_{out}) &=& \textbf{W}_{out} \textbf{r}.
\end{eqnarray}
Integrating these equations yields a prediction for a given $\rinit$ and $\Wout$, and is the function $R(\Wout, \rinit; \textbf{W}$, $\textbf{W}_{in}$, $\textbf{b})$ used in Algorithm 1.  We note that Eq. 3 has the same form as Eq. \ref{eqnarray:dynamics} from Sec. \ref{sec:definitions}, with the state $\textbf{x}$ replaced with the reservoir state $\textbf{r}$, the observation $\textbf{y}$ replaced with the output $\textbf{o}$, the parameter vector $\textbf{v}$ replaced with the parameter matrix $\textbf{W}_{out}$, and the initial state $\textbf{x}_0$ replaced with the initial reservoir state $\textbf{r}_0$. Thus, it is here that we see that a single, fixed RC with a distribution $p(\textbf{r}_0, \textbf{W}_{out})$ of initial conditions and parameters is capable of representing a distribution of time-series generated from $p(\textbf{x}_0, \textbf{v})$. We will exploit this connection in the next section to learn a latent representation of the underlying library parameters.

\subsection{Compressing RC Representations with Autoencoders }\label{subsec:autoencoders}

As described in Sec. \ref{sec:definitions}, a library member $\textbf{l}^i$ is determined by a particular parameter vector $\textbf{v}^i$ and initial condition $\textbf{x}_0^i$, so that we have $\textbf{l}^i = L(\textbf{x}_0^i, \textbf{v}^i)$.  Similarly, the training process $T$ identifies a map $(\rinit^i, \Wout^i)=T(\textbf{l}^i)$ from library members to the associated reservoir initial condition $\rinit^i$ and output weight matrix $\Wout^i$; this tuple of vectors is the \textit{RC features} that define a predictive RC model. In particular, we see that the composition of dynamics and training yield a relationship between the parameters and initial conditions of the underlying system to the corresponding RC features. That is, we have
\begin{equation}\label{eq:maps}
    (\textbf{r}_0, \textbf{W}_{out}) = T \circ L (\textbf{x}_0, \textbf{v}).
\end{equation}
For a typical application, the number of reservoir nodes $N$ can be on the order of thousands or more. If $m$ is the dimension of the desired outputs, then the dimension of the RC feature space is $(m+1)N$ and can be extremely large. However, Eq. \ref{eq:maps} reveals that if a set of RCs is trained on a library, then the RC features that encode dynamics approximating time series in the class generated by Eq. \ref{eqnarray:dynamics} actually lie on a manifold of much smaller dimension $\dim(\textbf{x}_0) + \dim(\textbf{v})$ within the larger space. If this manifold can be identified, then the problem of training an RC to predict a new time series that is similar to the library becomes a much lower-dimensional problem, and therefore can be accomplished with much less data.

Of course, the function $T \circ L$ cannot be known without direct measurements of $(\textbf{x}_0, \textbf{v)}$ which we assume to be unknown. However, an approximate, latent representation of this function can be learned with deep autoencoders \cite{baldi2012autoencoders, hinton2006reducing}.  Mathematically, the autoencoder is learning an identity function that is the composition of an encoding function $\latent = E(\textbf{r}_0, \textbf{W}_{out})$ and a decoding function 
\begin{equation}\label{eq:decoding}
D(\latent) = (\hat{\textbf{r}}_0, \hat{\textbf{W}}_{out}) \approx (\rinit, \Wout)
\end{equation}
which maps latent variable vectors $\textbf{e}$ to RC feature space.  We refer to the appendix for details on the autoencoder architecture, training, and hyperparameter considerations for the numerical experiments presented in this work.

\subsection{Optimizing Latent Variable Space}\label{subsec:search}

Even though a reservoir in feedback mode defined by Eq. 3 with fixed input and recurrent matrices is capable of representing a wide range of dynamical systems depending on $\Wout$, latent variable vectors decoded by Eq. \ref{eq:decoding} yield RC features that correspond to RCs that generate signals similar to those in the library.  To use this restricted space of RCs to predict the future evolution of the test signal $\textbf{s}$, we need a way to identify which $\latent$ decodes to $(\hat{\textbf{r}}_0, \hat{\textbf{W}}_{out})$ that best matches the short observed signal $\textbf{s}$.

Recall that, as each $(\textbf{x}_0^i, \textbf{v}^i)$ generates a time-series $\textbf{l}^i$, each $(\rinit, \Wout)$ can be used to generate a reservoir time series over the same time points on which $\textbf{s}$ is defined, which we denote by $\hat{\textbf{s}} = R(\textbf{r}_0, \textbf{W}_{out})$. If we restrict our attention to RC features that are decoded latent variable vectors, then we have $\hat{\textbf{s}} = R \circ D(\textbf{e)}$. The final step for forming our predictive model of $\textbf{s}$ is to find the RC feature vectors by 

\begin{eqnarray} \label{eqnarray:optloss}
    \hat{\textbf{e}} &=& \text{argmin}_{\textbf{e}} |R \circ D(\textbf{e}) - \textbf{s}|^2 \\
    (\hat{\textbf{r}}_0, \hat{\textbf{W}}_{out}) &=& D(\hat{\textbf{e}}) \nonumber
\end{eqnarray}

Because $\dim(\textbf{e})$ is relatively small, Eq. \ref{eqnarray:optloss} can be efficiently minimized with numerical algorithms such as Nelder-Mead \cite{nelder1965simplex} or dual annealing \cite{xiang1997generalized}.

Finally, we note that the composition of the decoding process (Eq. 5) and the RC prediction process (Eq. 3) forms a map from latent variable space to a time-series that is parallel to the process by which library members are generated from Eq. 1. Thus, if the meta-learning process succeeds, the map $T \circ D: \textbf{e} \rightarrow \textbf{l}^i$ is a latent representation of the map $L: (\textbf{x}_0, \textbf{v}) \rightarrow \textbf{l}$. We will see explicitly that this is the case for a toy example in the next section.

\section{Numerical Results}\label{sec:results}

In this section, we apply MARC to a couple of example problems, both to demonstrate the efficacy of our approach as well as illustrate some concepts discussed in the preceding sections. The parameters of the library and the test signals as well as the RCs, autoencoder, and latent space optimizer can be found in the Appendix.

\subsection{Sinusoidal Regression}\label{subsec:sine}
We first return to the sinusoidal regression problem discussed in the introduction, examined before in various forms in the context of few-shot meta-learning. As in \cite{finn2017model} and \cite{wu2018meta}, we consider a library of sine waves with unit frequency and amplitudes and initial phases uniformly sampled from $U[0.1, 5.0]$ and $U[0, \pi]$, respectively. As in the other works, we consider the test signal to be defined on $-5 \leq t \leq 5$. However, we consider the library signals to be defined on the longer interval $-6 \leq t \leq  5$ to ensure that the RC models are valid on the interval $-5 \leq t \leq 5$--see Appendix for more details on this issue. The library $\textbf{L}$ contains 100 examples, sampled from the distribution of initial conditions. Each member $\textbf{l}^i$ contains a set of 1,000 equally-spaced observations. Each test signal $\textbf{s}$ contains 10 randomly-sampled observations of a different sine wave. Details on the RC and autoencoder training are summarized in the appendix.  

\begin{figure}[h]
\begin{center}
\includegraphics[width=1\textwidth]{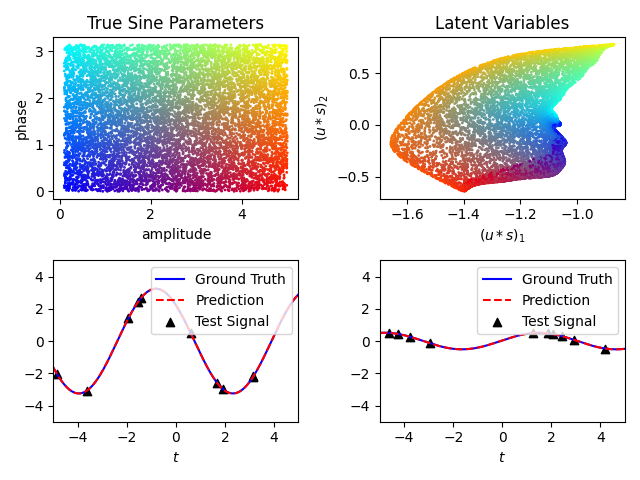}
\end{center}
\caption{Examination of the MARC algorithm applied to a toy regression problem. Upper left: 10,000 randomly sampled phase/amplitude points, plotted in the true 2-d parameter space, colored by location.  Upper right: The first two principal components of latent variable space, where the color of each point corresponds to its original phase and amplitude.   As we can see, the RC features map onto a smooth, compact manifold within latent variable space. Bottom: Two examples of the MARC algorithm applied to the sinusoidal regression problem. The blue curve represents the ground truth, while the dashed red line is the optimal reservoir time series $\hat{\textbf{s}}$. The 10 points, randomly sampled from the ground truth to form $\textbf{s}$, are indicated by black triangles.\label{fig:sineexample}}
\end{figure}

To evaluate the performance of the MARC algorithm, we report the mean squared error (MSE) over the complete sine wave in Table \ref{table:sine}, and visualize typical examples in Fig. \ref{fig:sineexample}.  In panels (a) and (b), we show how the true parameters (panel (a)) map to latent variable space (panel (b)).  In panels (c) and (d) we give two examples: from just 10 irregularly-spaced observations in $\textbf{s}$, the results of the MARC algorithm is clearly a sinusoidal wave that is a close approximation to the true data.

In Table \ref{table:sine}, we compare MARC to MAML \cite{finn2017model} and MeLA \cite{wu2018meta}, two generic metalearning algorithm proposals that report results for this task.  We find that MARC outperforms these algorithms by two orders of magnitude, but we note some caveats to this comparison.  In MARC, we use all available data (\textit{i.e.}, the entire sine wave sample for each library member) to train one RC per library member one time.  On the other hand, MAML and MeLA repeatedly sample only 10 observations from library members in the inner loop of their metalearning approach.  Because of the different usages of the library data, it is hard to directly compare the total data usage of MARC to MAML/MeLA.  Moreover, MAML/MeLA both use feedforward networks to perform nonlinear regression in this task, whereas we are using RCs for time series processing and interpret the sine wave as a time series.  As a result, these two approaches were applied to other deep learning architectures for other tasks, but the scope of this work is limited to time series prediction.  Finally, we note that MARC takes tens of minutes on a workstation-class laptop without GPU acceleration.

\begin{table}
\begin{center}
 \begin{tabular}{||c | c ||} 
 \hline
  & MSE \\ 
 \hline\
 MAML \cite{finn2017model} & 0.208 \\ 
 \hline
 MeLA \cite{wu2018meta} & 0.129 \\
 \hline
MARC (this work) & 0.0016 \\
 \hline
\end{tabular}
\end{center}
\caption{A comparison of the MSE in the sinusoidal regression problem for our algorithm and two state-of-the-art meta-learning approaches, as reported in \cite{wu2018meta}. \label{table:sine}
}
\end{table}

%To understand better what is happening, we visualize the action of $\latent = E \circ T \circ L(\xinit)$, \textit{i.e.}, the mapping from initial condition to latent variable space, in \ref{fig:sineexample}c-d.  We generate another 1,000 $\xinit$ (Fig. \ref{fig:sineexample}c), integrate Eq. \ref{eqnarray:sine} (the $L$ map),  train an RC on each (the $T$ map), and plot their mapped values in latent variable space (the $E$ map, Fig. \ref{fig:sineexample}d). We reiterate that these 1,000 members are purely for visualization purposes and not for training the autoencoder. As we see in Fig. \ref{fig:sineexample}a-b, the autoencoder has learned a latent representation of the trained RC features that is a smooth function of the sine wave parameters. We even see the expected periodicity, better illustrated in the cross section with fixed amplitude in Fig. \ref{Latent Space}c. Despite not having access to the internal degrees of freedom which define the library members and their corresponding RC models, the MARC algorithm has learned its notion of ``amplitude'' and ``phase.'' Important for the prospects of the MARC algorithm, the image of trained RC features in latent space appears to be compact and a smooth function of the true degrees of freedom that define $\textbf{L}$.

\subsection{Lorenz Prediction}\label{subsec:chaotic}

We now consider a more challenging example in the form of the Lorenz-63  \cite{lorenz1963deterministic} system.  The Lorenz-63 system is a three dimensional system of equations with three parameters; the long time behavior yields the famous ``butterfly attractor'' and it is a paradigmatic example of chaos used to benchmark time series prediction algorithms.

Chaotic systems, due to their fundamental property of exponentially divergent trajectories, are inherently difficult to predict, and any amount of error in the predictive model will eventually lead to uncorrelated predictions. Therefore, to quantify the performance of the MARC algorithm in this section, we report a valid prediction length $T_{valid}$, defined as the amount of time before the squared error first exceeds the variance of the target signal (details in the Appendix).
% , given explicitly by
% \begin{equation}
%     T_{valid} = min\{t-T_{close}; |\textbf{o}(t) - \textbf{x}(t)|^2 \geq var(\textbf{x}(t))\}
% \end{equation}
Intuitively, $T_{valid}$ defines the amount of time that the predictive model remains correlated with the true signal. It is best understood in units of the Lyapunov time $\Lambda_L$: small perturbations grow as $e^{t/\Lambda_L}$. The Lyapunov time of the Lorenz-63 system is $\Lambda_L \approx 1.104$s \cite{viswanath1998lyapunov}.

For this Lorenz-63 example, we construct a library of 100 example Lorenz attractors, where the parameters are drawn from a normal distribution around their canonical values, and each of the library members $\librarymember$ is a set of 5,000 observations separated by $\Delta t=0.01$, or, in terms of the Lyapunov time, about $50\Lambda_L$.   When trained with that much data, the baseline RCs that go into the autoencoder yield predictions with mean valid times approximately 2.87 $\Lambda_L$.  Rather than randomly sampling the temporal domain, the test signals $\textbf{s}$ consist of 10 data points separated by $\Delta t=0.01$, totalling approximately $0.1 \Lambda_L$. When trained with such limited data data, RCs do a poor job at both short-term forecasting (remaining valid for only 0.03 $\Lambda_L$ and at replicating the long-term behavior of the system. However, as seen in Fig. 3, the MARC algorithm yields a predictive model that is valid for more than a Lyapunov time and maintains the butterfly attractor of the Lorenz system. Over 100 test signals, the MARC prediction remains valid for an average of $1.06 \Lambda_L$, more than 10 times longer than the observed portion of the attractor.

%A typical trajectory is depicted in Fig. \ref{fig:lorenz}. for the nominal library and test parameters in Table \ref{Lorenz Hyperparameters}. As we can see, despite only a small segment of data from the unknown Lorenz system, MARC produces a model which is clearly Lorenz-like and tracks the target signal for multiple Lyapunov times. 

%In Fig. \ref{Lorenz Example}, we report the mean $T_{valid}$ as the difficulty parameters $|\textbf{L}|$, $|\textbf{s}$, and $\sigma$ are varied. We also compare against two RC baselines. The first baseline is an RC that has been trained in the conventional way on $\textbf{s}$ alone. As expected, this prediction performs quite poorly, due to the insufficiency in training data. The other baseline is an ``ideal'' RC that has been trained on a much longer $\textbf{s}$ that is not available to MARC algorithm. This baseline may be understood as the upper-bound of the performance of our meta-learning algorithm; it is ideal in the sense that the $\textbf{W}_{out}$ and $\textbf{r}_0$ are obtained from plenty of data. Contrary to expectations, MARC is relatively insensitive in this case to $|\textbf{L}|$ and $|\textbf{s}|$, although performance is degraded for increasing $\sigma$, as is expected. Some implications of these results for further research are discussed in Sec. \ref{Conclusions}.

\begin{figure}[h]
\begin{center}
\includegraphics[width=\textwidth]{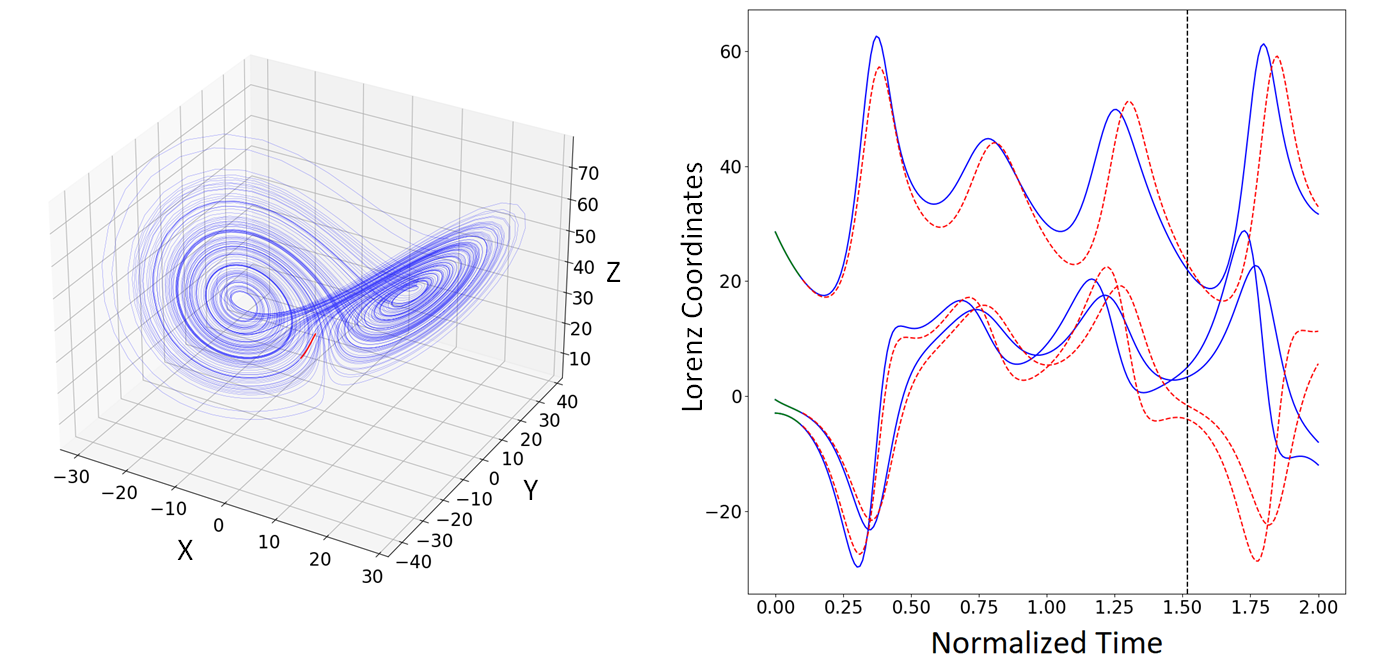}
\end{center}
\caption{An example of the MARC algorithm applied to the Lorenz system.  Left: An example Lorenz attractor (blue) generated by the result of the MARC approach using the 10 observed data points (red).   As can be seen, the decoded RC yields a textbook example, like decoded RCs in Sec. \ref{subsec:sine} all yield sines.  Right: The three components of the Lorenz system plotted vs. time illustrating the valid time calculation, both the true system (blue) and the MARC-based prediction (dashed red).  The 10 points, sequentially sampled from the ground truth to form $\textbf{s}$, are in green. The prediction remains valid for approximately 1.5 Lyapunov times, or until the dashed, vertical line, after which the systems become decorrelated due to the presence of chaos.\label{fig:lorenz}}
\end{figure}

% \begin{figure}[h]\label{Lorenz Parameter Sweep}
% \begin{center}
% \includegraphics[width=1\textwidth]{Figures/Lorenz Parameter Sweep.png}
% \end{center}
% \caption{A study of the end-to-end performance of the MARC algorithm applied to the Lorenz system for varying values of $|\textbf{L}|$, $|\textbf{s}|$, and $\sigma$, all of which are expected to increase the difficulty of the problem. Mean values and error bars are generated by running the experiment with 5 different instances of the complete algorithm, each tested against 25 different test signals.}
% \end{figure}

\section{Conclusions and Future Work}\label{sec:conclusions}
In this work, we have introduced a novel meta-learning approach to time series prediction from limited training data from the process of interest. The core idea behind our MARC method is that by using a flexible, general machine-learning model with a well-behaved training algorithm, the model features resulting from training a class of related processes lies on a compact, lower-dimensional manifold. The manifold can be effectively learned with standard autoencoders, and the latent variables can be searched to find a model that matches a small amount of test data.  The resulting model can be used for prediction and avoids overfitting, because there are only a few free parameters to adjust to new data.  We have demonstrated our approach by applying it to a simple regression example and obtaining results that greatly exceed state-of-the-art gradient-based techniques. We then apply it to a nontrivial example involving chaos, demonstrating the wide applicability of the technique.  We have explicitly explored some of the properties of latent variable space and given a precise notion to the scope of situations in which we expect our algorithm to apply.

There are several avenues to improve the results we have obtained. First, the most apparent difficulty with our approach is the need for ultra-high-precision autoencoders. For the results in this work, thousands of epochs were trained to obtain an autoencoder with MSE on the order of $10^{-4}$ or smaller. It is likely that better predictive models can be obtained if this error were to be even further reduced, due to the high sensitivity of the predicted signals $\hat{\textbf{s}}$ on errors in $\textbf{W}_{out}$ and $\textbf{r}_0$. On the other hand, the RC algorithm itself might be refined to be more robust to these errors, such that more easily obtained autoencoder performance would be satisfactory. This avenue is suggested by some preliminary analysis on the dependence of the MARC algorithm on the RC regularization parameter (see Appendix), which suggests that values which are much higher than optimal for conventional RC actually perform best in this scenario. This may be because increasing the penalty for variance in $\textbf{W}_{out}$ makes more error tolerable.

Second, the generic nature of the optimization routine in Sec. \ref{subsec:autoencoders} can be improved by further study.  Nelder-Mead is a Jacobian-free method chosen mostly for expedience, but given that the error $\hat{\textbf{s}}-\textbf{s}$ is the function of a recurrent neural network (the RC) and a feedforward neural network (the autoencoder), we could explicitly compute the Jacobian and use other optimization schemes to more efficiently identify the correct latent variable $\textbf{e}$.

Thirdly, the issue of choosing an appropriate latent variable dimension is an issue.  In the presented model system examples, we knew the proper dimensionality of the manifold; in practice with real data, this dimension is unknown.   In our numerical experiments, we used a slightly larger latent variable dimension than strictly necessary to accomodate loss due to the autoencoder (see Appendix), but a principled approach to select this crucial hyperparameter is needed before widespread practical application.  Choosing a dimension that is too large greatly increases the burden on the optimizer so that the probability of getting stuck in a local optimium increases, while choosing a dimension that is too small leads to  a lack of expressiveness in the RC model that may miss important dynamics.

Notwithstanding these issues, we are optimistic that the ideas behind MARC can have far-reaching application beyond the focus of time series prediction presented in this work.  Firstly, it may be the case that RCs trained for other tasks (such as control \cite{canaday2021, antonelo2014learning}, signal classification \cite{escalona2014electrocardiogram, tanaka2017waveform}, and anomaly detection \cite{obst2008using, chen2020imbalanced}) or through  reinforcement learning \cite{Chang2020RL} can be similarly compressed with autoencoders so that new, related systems can be controlled or classified with greatly reduced training data requirements. Secondly, we have focused here on RCs--both because of our initial focus on time series prediction and due to attractive numerical properties--but the general scheme of Sec. \ref{sec:algorithm} may be applied to other to machine learning models, such as deep network architectures, so that other tasks such as computer vision may benefit from this approach.  Finally, the ideas behind MARC may be a tool in scientific discovery: systems with different dynamics cluster in latent variable space, a behavior which may be useful in exploratory data analysis.  Furthermore, in experimental cases where data collection is hard and/or expensive, identifying regions of latent variable space that are sparsely populated may be useful to guide future data collection.  Finally, RCs have been proposed as a computationally-efficient method to replace integration of stiff ODEs \cite{mattheakis2021unsupervised, anantharaman2020accelerating}; combining these ideas with a metalearning parameter interpolation scheme could greatly reduce the computational requirements of scientific computing.  

% Commented the acknowledgement section for blind peer review
\section*{Acknowledgements}
This work was supported by DARPA under contract W31P4Q-20-C-0077.  We thank Edward Ott, Brian Hunt, and Jiangying Zhou for useful comments and discussion.  The views, opinions, and/or findings expressed are those of the author(s) and should not be interpreted as representing the official views or policies of the Department of Defense or the U.S. Government.

\bibliography{references}
\bibliographystyle{unsrt}

\appendix

\section{Reservoir Computing}\label{Reservoir Computing}
In this appendix, we provide additional details for our implementation of RC in the numerical examples in this work, including hyperparameter considerations for the specific experiments. Recall from Sec. 3.1 that the first step in the MARC algorithm is to learn independent predictive models for each available time series. Thus, we require a flexible machine learning framework that is capable of representing a range of dynamical systems. Further, we require the training algorithm to produce model features that are smooth, deterministic functions of the training data such that Eq. 4 holds. A commonly-used type of RC known as an echo-state network (ESN) (\cite{jaeger2001echo}) satisfies these requirements and is used in this work, though we note that other machine learning frameworks may be suitable as well.

\subsection{Echo State Networks}\label{Echo State Networks}
In the ESN formulation of RC, the reservoir is a recurrent neural network model whose state $\textbf{r}(t)$ is a column vector with $\textbf{r}^T(t) = [r_1(t), r_2(t), ..., r_N(t)]$ where the scalar state of each network node $l$ is $r_l(t)$ and the reservoir dynamics are defined by Eq. \ref{Reservoir Dynamics}.   In the equations, $c$ is a time constant, $\textbf{W}$ is a matrix of recurrent connections, $\textbf{W}_{in}$ is an input-connectivity matrix, $\textbf{b}$ is a bias vector, and $\textbf{W}_{out}$ is the trained output matrix. Importantly, we emphasize that the network dynamics, \textit{i.e.}, $c$, $\textbf{W}$, $\textbf{W}_{in}$, and $\textbf{b}$ are held constant during training. Note also that the trainable weights $\textbf{W}_{out}$ do not impact the internal reservoir dynamics and, therefore, can be trained after observing the reservoir response $\textbf{r}(t)$ to the input signal $\textbf{u}(t)$.

There are a number of different possibilities for constructing the the network-defining matrices $\textbf{W}$, $\textbf{W}_{in}$, and $\textbf{b}$. In this work, we do so according to the following procedures. We choose elements of the connectivity matrix $\textbf{W}$ to be nonzero with probability $k/N$, where $N$ is the number of network nodes and $k$ is the mean in-degree of the network. Nonzero elements are chosen from a normal distribution with zero mean and an arbitrary variance. The elements of $\textbf{W}$ are then rescaled to fix the spectral radius to $\rho_\textbf{W}$. This scaling step is important to ensure that the dynamics of $\textbf{r}(t)$ ``forget'' its initial condition and previous inputs and to help match the time scale of the RC dynamics to the input signal \cite{jaeger2001echo}. The matrix $\textbf{W}_{in}$ and the vector $\textbf{b}$ and have elements drawn from a symmetric uniform distribution with maximum value $\sigma_{in}$ and $\sigma_\textbf{b}$, respectively. The former hyperparameter determines the coupling strength of the input to the reservoir, while the latter promotes a diversity of individual node dynamics. 

Generally speaking, $c$ describes the dominant time-scale of the reservoir dynamics. Since the RC is used in this work to model a time series, $c$ should be chosen to roughly match the dominant time-scale $\textbf{u}(t)$. Because $\sigma_{in}$ scales the input signal, it is generally taken to be approximately the inverse of the scale (standard deviation or maximum absolute value) of the input signal. Finally, $\sigma_\textbf{b}$ is chosen between 0 and 1.

 \subsection{Training}\label{ESN Training}

 For a review of RC methods, we refer the reader to \cite{lukovsevivcius2012reservoir}.  As outlined in the reference, RC training is most often achieved by linear regression with Tikhonov regularization \cite{tikhonov2013numerical}. Let $\textbf{R}$ be the matrix of discrete observations of the reservoir state space. That is, the $n^{th}$ column of $\textbf{R}$ is $\textbf{r}(T_{init}+n\Delta t)$, where $T_{init}$ is the warm-up time, $\Delta t$ is the sampling period, and $n$ is an integer. Similarly, let $\textbf{O}^d$ be a matrix whose $n^{th}$ column is the desired output at time $(T_{init}+n\Delta t)$. Then the output matrix $\textbf{W}_{out}$ is chosen to minimize the loss given by

\begin{equation}\label{Training Loss}
     |\textbf{W}_{out}\textbf{R} - \textbf{O}^d|^2 + \alpha | \textbf{W}_{out}|^2,
 \end{equation}
 where $|\textbf{M}|^2$ for a rectangular matrix $\textbf{M}$ is defined as the trace of $\textbf{MM}^T$ and $\alpha$ is a small, positive regularization parameter that prevents overfitting. The regularization $\alpha$ is commonly selected with cross-validation techniques. However, we find that the MARC algorithm benefits from a significantly larger value of $\alpha$ than what is optimal for using RC to predict a single time series. (This may be because less sensitive $\textbf{W}_{out}$ values produce a manifold that is smoother and easier for the autoencoder to learn.)

Conveniently, the minimum to Eq. \ref{Training Loss} can be expressed in the closed form given by

\begin{equation}\label{Training Loss Explicit}
     \textbf{W}_{out} = (\textbf{R}^T \textbf{R} + \alpha \textbf{I})^{-1} \textbf{R}^T \textbf{O}^d.
\end{equation}
As the right-hand-side of Eq. \ref{Training Loss Explicit} is the composition of continuous functions, this makes explicit the claim in the main text that the trained RC features are continuous functions of the training data.
 
The RC hyperparameters for the sine-regression and Lorenz-63 experiments are presented in Tab. 2 and 3, respectively.

\section{RC Compression with Autoencoders }\label{sec:autoencoders}

To obtain a mapping from a low-dimensional latent space to the continuous set of RC features, we compress the features through an autoencoder with a low-dimensional intermediate layer. Thus, the autoencoder $A$ is the composition of an encoding function $E$ and a decoding function $D$, only the latter of which is used for fitting RC features to a new test signal. A schematic of this architecture is presented in Fig. 6.

For the examples in this paper, we construct the autoencoder with five hidden layers, the middle one being the smallest size and identified as $\textbf{e}$. More specifically, the autoencoder contains an input layer of size $(m+1)N$, followed by two hidden layer, followed by the latent layer of size $dim(\textbf{e})$, followed by two more hidden layer and finally an output layer of size $(m+1)N$. The first six layers have sigmoid activation functions, while the last layer is linear. All of the layers are dense and have no intra-layer connections. The network is trained using the ADAM algorithm \cite{chilimbi2014project} with learning rate fixed at 0.0001. We employ an early-stopping procedure \cite{caruana2001overfitting} to avoid over-fitting, where 10\% of the training samples are reserved for validation, and training ends when the validation error worsens or does not improve for a number of epochs. For the sine-regression experiment, the successive hidden layers are of dimension 200, 200, 4, 200, 200, respectively. For the Lorenz-63 experiment, these dimensions are 600, 200, 7, 200, 600, respectively.

\begin{figure}[h]\label{Autoencoder}
 \begin{center}
 \includegraphics[width=0.68\textwidth]{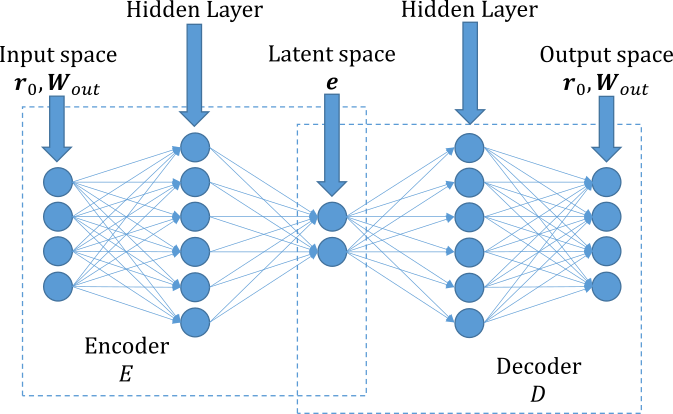}
 \end{center}
 \caption{The autoencoder architecture. An autoencoder takes $(\textbf{r}_0, \textbf{W}_{out})$ as input and produces an approximation to $(\textbf{r}_0, \textbf{W}_{out})$ as its output. The decoder layer approximates $(\textbf{r}_0, \textbf{W}_{out})$ from a small-dimensional latent space variable $\textbf{e}$, mimicking the process of generating $(\textbf{r}_0, \textbf{W}_{out})$ from $(\textbf{x}_0, \textbf{v})$.}
 \end{figure}

 \section{Optimizing in Latent Variable Space}\label{Optimizing in Latent Variable Space}

Recall that, as each $(\textbf{x}_0^i, \textbf{v}^i)$ generates a time-series $l^i$, each $(\textbf{r}_0, \textbf{W}_{out})$ can be used to generate a reservoir time series over the same time points on which $s$ is defined, which we denote by $\hat{\textbf{s}} = R(\textbf{r}_0, \textbf{W}_{out})$. If we restrict our attention to RC features that are decoded latent variable vectors, then we have $\hat{\textbf{s}} = R \circ D(\textbf{e)}$. The final step for forming our predictive model of $\textbf{s}$ is to find the RC feature vectors by minimizing the loss in Eq. 6.

Because $dim(\textbf{e})$ is small, Eq. 6 can be efficiently minimized with a variety of numerical algorithms. For this purpose, we use an efficient global search algorithm based on simulated dual annealing \cite{xiang1997generalized}. For alle xperiments in this paper, we implement this algorithm with a maximum iteration limit of 100 and a local search based on the Nelder-Mead method \cite{nelder1965simplex}.

 \begin{table}\label{Sinusoidal Hyperparameters}
 \begin{center}
  \begin{tabular}{||c c | c c | c c ||} 
 \hline
  \shortstack{library, test \\ parameter} & value & \shortstack{RC \\ parameter} & value & \shortstack{autoencoder \\ parameter} & value \\ [0.5ex] 
  \hline\hline
  $|\textbf{L}|$ & 100 & $N$ & 1,000 & $dim(\textbf{h}_1)$ & 200 \\ 
  \hline
  $|\textbf{l}^i|$ & 1,000 & $k$ & 100 & $dim(\textbf{h}_2)$ & 200 \\
  \hline
  $|\textbf{s}|$ & 10 & $\rho_\textbf{W}$ & 1.0 & $dim(\textbf{e})$ & 4 \\
  \hline
   & & $\sigma_{in}$ & 1.0 & & \\
  \hline
   & & $\sigma_\textbf{b}$ & 1.5 & & \\
  \hline
   & & $c$ & $1.0$ & & \\
  \hline
   & & $\alpha$ & $5e^{-6}$ & & \\
 \hline
   & & $T_{init}$ & 1.0 & & \\
  \hline
 \end{tabular}
 \end{center}
 \caption{The hyperparameters for the MARC algorithm used for the sinusoidal regression experiments described in Sec. \ref{subsec:sine}. The same hyperparameters (except for $dim(\textbf{e})$) are used for the multi-modal experiments described in Appendix E.}
 \end{table}

 \begin{table}\label{Lorenz Hyperparameters}
  \begin{center}
  \begin{tabular}{||c c | c c | c c ||}
  \hline
  \shortstack{library, test \\ parameter} & value & \shortstack{RC \\ parameter} & value & \shortstack{autoencoder \\ parameter} & value \\ [0.5ex] 
  \hline\hline
  $|\textbf{L}|$ & 1,000 & $N$ & 1,000 & $dim(\textbf{h}_1)$ & 600 \\ 
  \hline
  $|\textbf{l}^i|$ & 5,000 & $k$ & 100 & $dim(\textbf{h}_2)$ & 200\\
  \hline
  $|\textbf{s}|$ & 10 & $\rho_\textbf{W}$ & 0.8 & $dim(\textbf{e})$ & 7 \\
  \hline
    & & $\sigma_{in}$ & 0.05 & & \\
  \hline
   & & $\sigma_\textbf{b}$ & 0.5 & & \\
  \hline
   & & $c$ & $1.0$ & & \\
  \hline
   & & $\alpha$ & $5e^{-4}$  & & \\
  \hline
   & & $T_{init}$ & 10.0 & & \\
  \hline
 \end{tabular}
 \end{center}
 \caption{The hyperparameters for the MARC algorithm used for the Lorenz prediction experiments described in Sec. \ref{subsec:chaotic}.}
 \end{table}

\section{Details on the Lorenz Experiment}

In Sec. \ref{subsec:chaotic}, we discussed the challenging problem of forecasting the Lorenz-63 system from limited data. The Lorenz-63 system is described by the following dynamical equations:
 \begin{eqnarray}
      \dot{x}_1 &=& v_1(x_2-x_1) \nonumber \\
     \dot{x}_2 &=& x_1(v_2-x_3) - x_2 \\ 
    \dot{x}_3 &=& x_2x_2 - v_3x_3 \nonumber \\
    \textbf{y} = \textbf{x}.
\end{eqnarray}
The parameters $v_1$, $v_2$, and $v_3$ are sampled from the uniform distributions $U[10, 15]$, $U[28, 42]$, and $U[8/3, 12/3]$, respectively. Initial conditions are randomly sampled from the attractor by allowing the system to propagate for $100\Lambda_L$ before recording the time series. The hyperparameters used for the RCs for the numerical results in Sec. \ref{subsec:chaotic} are presented in Table 3.

\section{Applicability to Multi-modal Tasks}\label{Multimodal}
In this section, we discuss the applicability of the MARC algorithm to multi-modal tasks where the library $\textbf{L}$ consists of different types of functions. In particular, we consider the experiment from \cite{yao2019hierarchically} where each $\textbf{l}^i$ and $\textbf{s}$ are generated, with equal probability from the following four functions:
\begin{eqnarray}
    y(x) = A \space\text{sin}(\omega x + b) \\
    y(x) = A_l x + B_l \\
    y(x) = A_q x^2 + B_q x + C_q \\
    y(x) = A_c x^3 + B_c x^2 + C_c x + D_c
\end{eqnarray},
where the coefficients are drawn from the distributions $A \leftarrow U[0.1, 0.5]$, $\omega \leftarrow U[0.8, 1.2]$, $b \leftarrow U[0, 2\pi]$, $A_l \leftarrow U[0.1, 0.5]$, $B_l \leftarrow U[0.1, 0.5]$, $A_q \leftarrow U[0.1, 0.5]$, $B_q \leftarrow U[0.1, 0.5]$, $C_q \leftarrow U[0.1, 0.5]$, $A_c \leftarrow U[0.1, 0.5]$, $B_c \leftarrow U[0.1, 0.5]$, $C_c \leftarrow U[0.1, 0.5]$, and $D_c \leftarrow U[0.1, 0.5]$.

We train the MARC algorithm on a library of size $\textbf{L}=1,000$. The latent dimension $dim(\textbf{e})$ is increased to 7 to account for the increased degrees of freedom in the library, but hyperparameters are otherwise as in the sine regression experiments. Across 800 test signals, we find the mean MSE (0.394) to be comparable to results reported in \cite{yao2019hierarchically}, being exceeded only by methods that are designed with multi-modal distributions in mind.

\begin{figure}[h]\label{Sinesandfriends Examples}
 \begin{center}
 \includegraphics[width=0.95\textwidth]{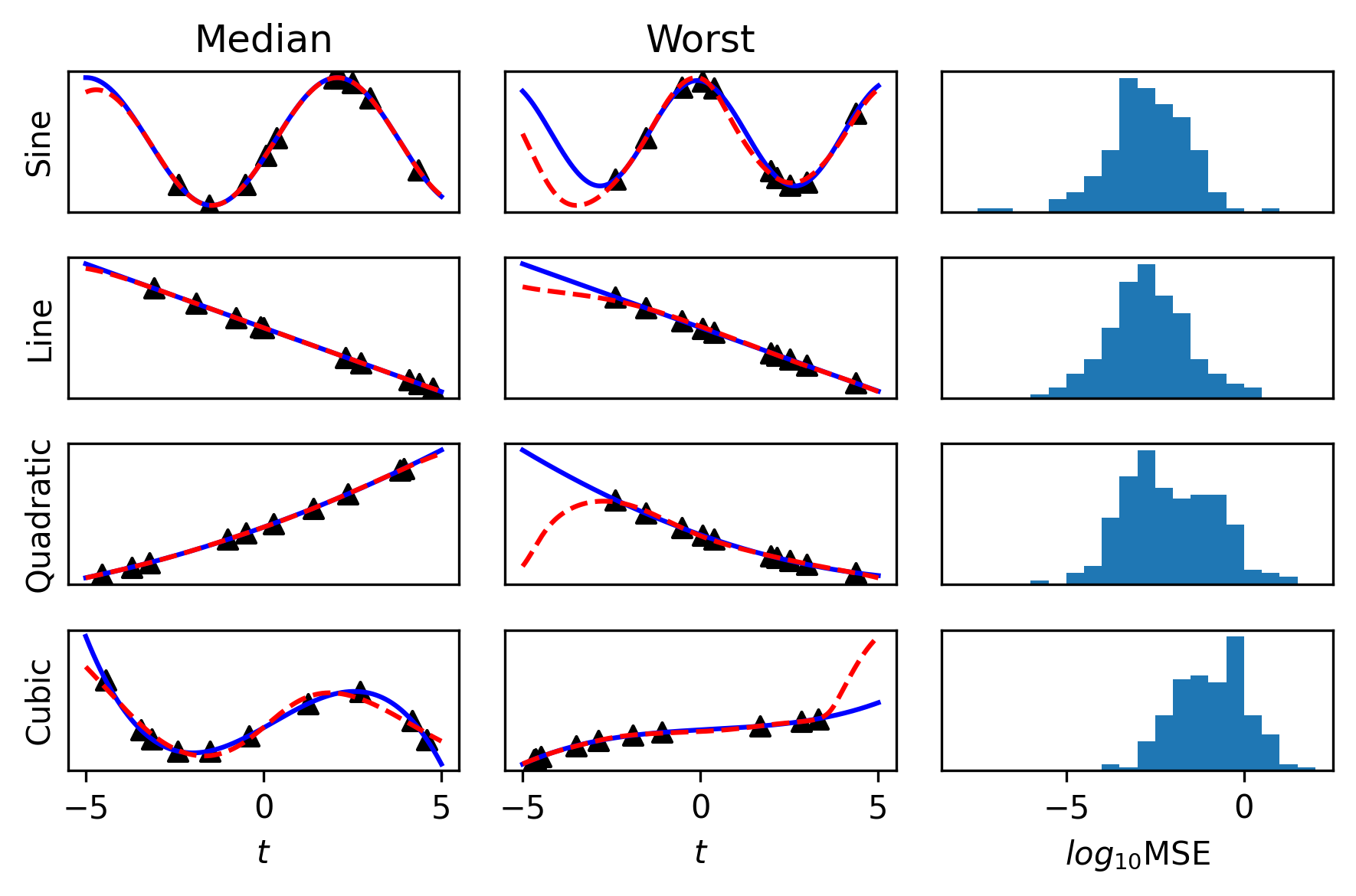}
 \end{center}
 \caption{Illustrative examples of the MARC algorithm applied to the multi-modal example. The first, second, third, and fourth rows correspond to the sine, line, quadratic, and cubic test signals, respectively. The first (second) column depicts the test with the median (worst) MSE out of 800 total experiments. Black triangles indicate the observed points in $\textbf{s}$. Blue lines indicate the ground truth, while dashed red lines indicate the MARC prediction. The third column contains the distribution of errors in logarithmic space.}
 \end{figure}

Upon closer inspection of MARC results on this toy problem, we see that the mean error is dominated by a few outliers among the many test signals. In fact, the \textit{median} MSE is several orders of magnitude better at 0.006, and that errors appear to be normally distributed in logarithmic space, as illustrated in Fig. \ref{Sinesandfriends Examples}. Further depicted in Fig. \ref{Sinesandfriends Examples} are the median and worst-case performances on the test signals. We see here that the median performance is quite good, while worst-case scenarios are driven by predictions that extrapolate poorly and appear to lose the the sinusoidal or polynomial character of the library members. We also see that, generally, the cubic test signals are most difficult for MARC to learn simultaneously with the sinusoidal, linear, and quadratic examples.

We further investigate the source of poor predictions in Fig. xx where we plot the relationship between the optimizer loss (the last step of MARC) and the final MSE. It is clear that poor performance is predominantly a result of not finding an adequate global minimum solution to Eq. 6.

The above observations, together with the fact that the first two steps (RC training and autoencoding) of MARC perform exceptionally well on this problem (as measured by autonomous RC prediction error and autoencoding error, respectively), suggest that the difficulty of applying MARC to the multi-modal problem arises from the optimization step. This is perhaps due to an increase in local minima to Eq. 6. This example further motivates a more careful consideration of latent-space optimization that is tuned to behave well with the MARC algorithm. It also suggests possible refinements to MARC, such as adding a term to the loss in Eq. that penalizes deviation from the library-created manifold, or a hierarchical structure to the latent space in the spirit of HSML. These ideas are the subject of ongoing research.

\begin{figure}[h]\label{Optimizer Loss versus MSE}
 \begin{center}
 \includegraphics[width=0.75\textwidth]{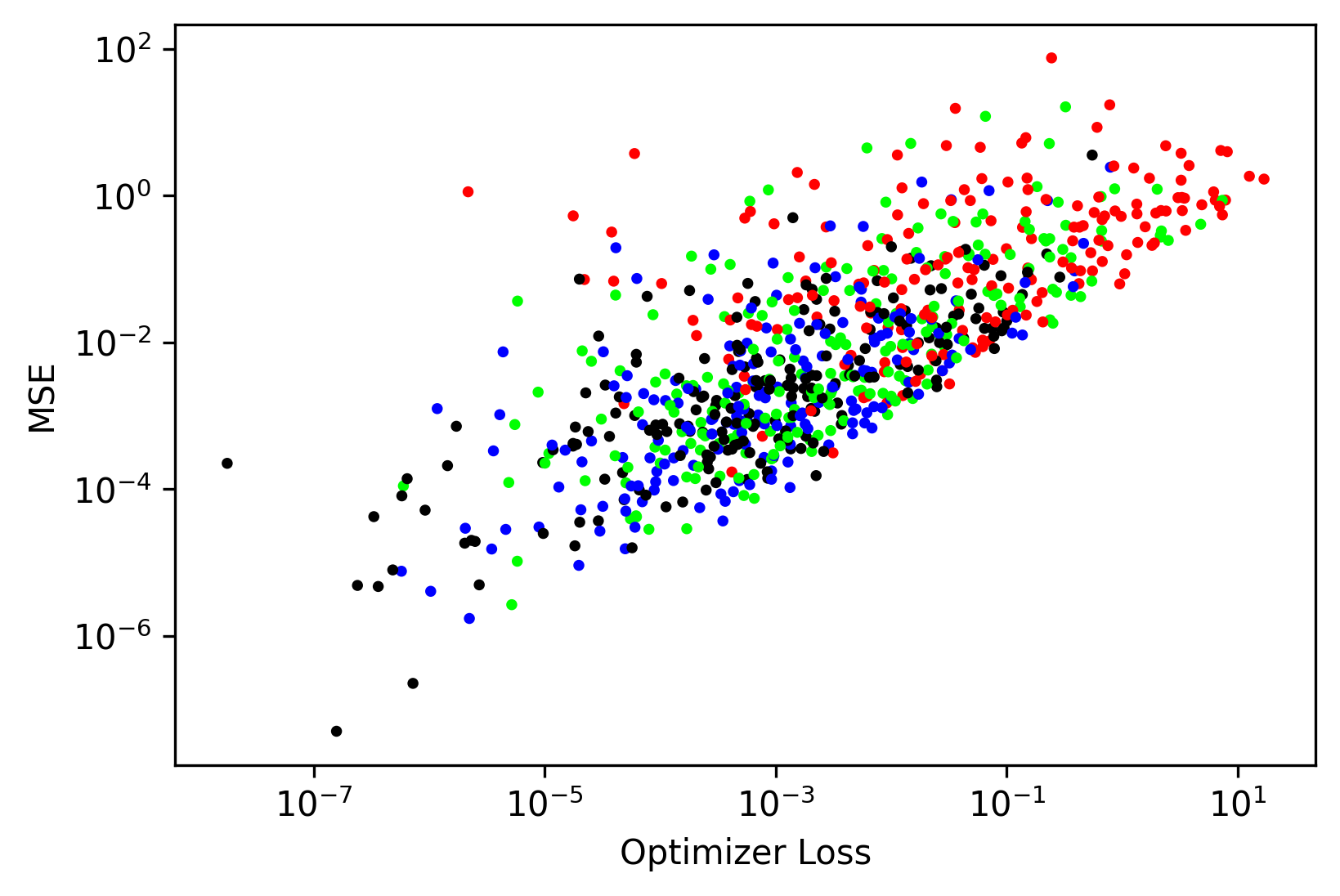}
 \end{center}
 \caption{Optimizer loss versus final MSE for the multi-modal experiments. Black, blue, green, and red dots correspond to sines, lines, quadratic functions, and cubic functions, respectively.}
 \end{figure}

\end{document}